\documentclass{INTERSPEECH2023}

\interspeechcameraready 

\usepackage[mathscr]{euscript}
\usepackage{amsmath}
\usepackage{multirow}
\usepackage{xcolor}
\usepackage{pgfplots}
\usepackage{tikz}
\usepackage{xcolor}

\title{Multilingual Contextual Adapters To Improve Custom Word Recognition In Low-resource Languages}
\name{Devang Kulshreshtha$^*$\quad Saket Dingliwal$^*$\quad Brady Houston\quad Sravan Bodapati}
\address{AWS AI Labs}
\email{\{kulshrde, skdin, hstbrady, sravanb\}@amazon.com}
\begin{document}

\maketitle
\def\thefootnote{*}\footnotetext{Equal contribution}\def\thefootnote{\arabic{footnote}}
\begin{abstract}
     Connectionist Temporal Classification (CTC) models are popular for their balance between speed and performance for Automatic Speech Recognition (ASR). However, these CTC models still struggle in other areas, such as personalization towards custom words. A recent approach explores Contextual Adapters, wherein an attention-based biasing model for CTC is used to improve the recognition of custom entities. While this approach works well with enough data, we showcase that it isn't an effective strategy for low-resource languages. In this work, we propose a supervision loss for smoother training of the Contextual Adapters. Further, we explore a multilingual strategy to improve performance with limited training data. Our method achieves 48\% F1 improvement in retrieving unseen custom entities for a low-resource language. Interestingly, as a by-product of training the Contextual Adapters, we see a 5-11\% Word Error Rate (WER) reduction in the performance of the base CTC model as well.
     

\end{abstract}

\section{Introduction}
\label{sec:intro}
 End-to-End (E2E) models still stubbornly lag their hybrid predecessors in custom entity recognition \cite{sainath2018no, bruguier2016learning}, where a model is presented with a list of custom words and is expected to better recognize them. Hybrid models’ modularity and explicit frame-wise phonetic targets are naturally amenable to various boosting approaches used for this task. However, recent work addresses this problem for E2E models by exploring various methods to boost custom words, such as LM fusion \cite{le2021deep, kannan2018analysis}, multi-modal training \cite{gourav2021personalization} or modification of the E2E model itself \cite{das2022listen, sathyendra2022contextual, pundak2018deep, bruguier2019phoebe}.
 
CTC-based E2E models \cite{salazar2019self}, in particular, are a challenge for current boosting methods because of its time-synchronous and conditionally independent outputs. Recently, a two-phase boosting approach for CTC explores improving its customization \cite{dingliwal2023personalization}. In this work, Contextual Adapters are trained to attend to custom entities using representations from different layers of an already trained acoustic to subword encoder and copy them to the output when they were present in speech. Together with an on-the-fly boosting approach during decoding, this approach resulted in improved recognition of out-of-vocabulary words (OOVs) on English VoxPopuli dataset by 60\% (F1-score) with minimal impact on overall WER. 

However, the dependency of Contextual Adapters’ ability to learn to boost patterns on data volume remains unclear. We identify that this ability significantly reduces for low-resource languages. We attribute our finding to three main reasons: (1) the training objective for the Contextual Adapters lacks direct guidance and is hard to optimize, specially in a low-data setting (2) the CTC encoder trained using limited data produces low-quality audio representations (3) low-resource languages may not have enough diversity in the training data for the Contextual Adapters to copy and boost from the weak encoder outputs. 

In this work, we systematically explore different strategies to mitigate the above issues. First, we propose a novel multitask cross-entropy loss for training the Contextual Adapters, which provides additional supervision to copy the right entity word whenever required. Second, we leverage multilingual training of E2E models, which has been shown to significantly improve the performance for low-resource languages in terms of overall WER \cite{Pratap2020, Cho2019, 8461802}. Lastly, we expand multilingual training to include Contextual Adapters to bootstrap the learning of copying and boosting mechanism for a low-resource language. Our proposed training strategy results in models that exhibit robust overall performance and can be customized to any list of entities. Our model, that uses just 100 hours of Portuguese language for training, achieves a 48\% absolute improvement of F1 scores in the retrieval of custom words over monolingual training of Contextual Adapters. Interestingly, we find that fine-tuning the multilingual encoder and multilingual Contextual Adapters together on monolingual data reduces WER by 5-11\% over fine-tuning the multilingual encoder alone for low-resource languages. 


\section{Background}
\label{subsec:background_adapters}
\begin{figure*}
    \centering
    \includegraphics[width=0.75\linewidth]{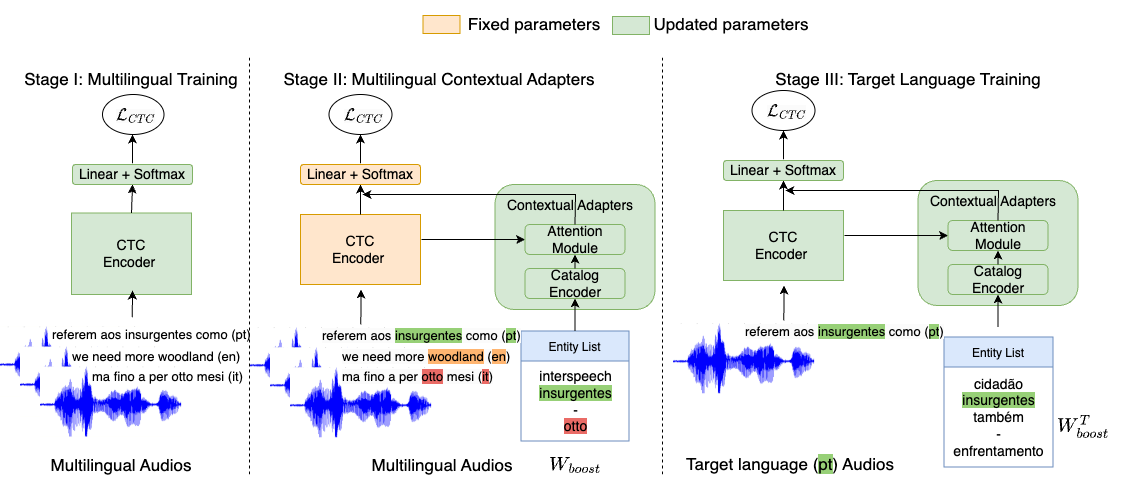}
    \caption{Three stages for training multilingual contextual adapters (Track "b"). Stage I trains a multilingual encoder, Stage II learns multilingual Contextual Adapters by freezing the encoder, while Stage III jointly optimizes both the components on the target language. }
    \vspace{-0.4cm}
    \label{fig:arch}
\end{figure*}


\noindent \textbf{Contextual Adapters for CTC Encoder}: As showcased in the Figure \ref{fig:arch}, a CTC encoder (parameterized by $\theta_{ctc}$) takes in an audio as input, passes it through multiple Conformer \cite{gulati2020conformer} blocks to produce a sequence of word piece posteriors, which is then optimized using the CTC loss ($\mathcal{L}_{CTC}$) \cite{graves2006connectionist}.
One of the methods to boost a list of custom words includes training a separate attention-based module called Contextual Adapters (parameterized by $\theta_{CA}$) by freezing $\theta_{ctc}$ and optimizing $\mathcal{L}_{CTC}$ on a subset ($D^{\text{boost}}$) of the training data that contains at least one entity/rare term in the transcript. 
A training sample $d = (x,y,W^{boost})$ for the Contextual Adapters contains an audio $x$ with $T$ frames, transcript $y$ and a list of $K$ random entity words $W^{boost}$. Let $w^{boost}$ be an entity word present in $y$, then $w^{boost} \in W^{boost}$. As highlighted in the Figure \ref{fig:arch}, the Contextual Adapters is comprised of two main components: 1) an LSTM catalog encoder, which encodes the sub-word tokens comprising each entity term in the list $W^{boost}$ and 2) an attention-based biasing adapter, which attends over the union of each word in the list $W^{boost}$ and a special $\langle nb \rangle$ (no-bias) token to learn a context vector. This context vector is added to the final encoder layer output of the base CTC model and the sum is then passed into the softmax layer. The attention focuses on the $\langle nb \rangle$ token for the time frames which do not correspond to any word in the list, while it attends to the sub-words corresponding to $w^{boost}$ for the frames in which it is spoken. After the module identifies the right word to boost, the probability of the subword sequence corresponding to that word is increased in the output. This learned `copying` mechanism can then be used to personalize an ASR system to a custom entity list during inference.\\

\noindent\textbf{Multilingual End-to-end Model Training}: Multilingual modeling has seen a resurgence in parallel with E2E modeling. As E2E models do not require explicit alignments nor phonetic targets, the data from individual languages can be pooled together using a common sub-word vocabulary \cite{Toshniwal2018}. When used in low-resource language settings, this simple approach yields impressive improvements over monolingual models \cite{Pratap2020, Cho2019, 8461802}. These pooled models can be further fine-tuned on individual languages to improve performance \cite{houston2023exploration}. 



\section{Methodology}

\subsection{Supervision Loss for Contextual Adapters}\label{subsec:adapter_loss}
Prior work used various approaches for training attention-based modules for personalization, such as a curriculum learning-based approach that slowly increases the size of the entity word list i.e. $K$ during training \cite{dingliwal2023personalization}, adding a special suffix to each entity word in the ground truth \cite{pundak2018deep}, and using separate labeled data for training the Contextual Adapters and the CTC model \cite{sathyendra2022contextual}. However, they optimize $\mathcal{L}_{ctc}$ for training the attention module’s parameters. This loss does not explicitly capture the problem of choosing the right entity word and boosting the corresponding sub-word sequence. 
In this work, we introduce an adapter-specific loss function to train the Contextual Adapters, which is added to the CTC loss, and then jointly optimized as a multi-task objective. For a training sample as defined in Section \ref{subsec:background_adapters}, at any time-step $t \in T$, attention weight of any entity word $w_k, k \in K$ in the Contextual Adapters is the normalized dot product between the encoder output at that time step $t$ (query) and the catalog encoder output for $w_k$ (key). Let $\mathcal{W}_{\langle nb \rangle}^{t}, \mathcal{W}_{k}^{t}$ be the attention weight given to $\langle nb \rangle$ token and the word $w_k$ at time step $t$ respectively. We use these attention weights to directly predict the right entity word from the list. We add a $K$-class cross-entropy loss (CE loss) as defined in Eq. \ref{eqn:ce}. We only want to add this loss for those time-frames in the input audio where the entity word was spoken. Therefore, for all other audio frames, we maximize the attention given to the $\langle nb \rangle$ token, and hence, we weigh the time frames in an inverse proportion to $\mathcal{W}_{\langle nb \rangle}$
\vspace{-0.3cm}
\begin{equation}\label{eqn:ce}
    \mathcal{L}_{CE} (.) = - \sum_{t=1}^{T} (1 - \mathcal{W}_{\langle nb \rangle}^{t} ) \log (\sigma (\mathcal{W}_{k’}^t))
\vspace{-0.2cm}
\end{equation}
where $k’$ is the index of the correct word i.e. $w_{k’} = w^{boost}$ and $\sigma(.)$ is the Softmax function over all the attention weights $\mathcal{W}_k^t, k \in K$. 
Finally, for training our Contextual Adapters with limited paired text and audio data, we  minimize a weighted combination of the two losses i.e. $\mathcal{L}_{net} = \mathcal{L}_{CTC} + \alpha \mathcal{L}_{CE}$.

\subsection{Multilingual Contextual Adapters}
For low-resource languages, training Contextual Adapters for custom word boosting is challenging as 1) a weak CTC model outputs sub-optimal audio embeddings, preventing the Contextual Adapters to identify on the right word, and 2) learning the parameters for the Contextual Adapters requires a large volume of labeled data. We add multilingual training data to both CTC and the  Contextual Adapters helping in alleviating both the above issues. We investigate two different tracks for multilingual training with three stages for each track. For both these tracks, we first train a baseline CTC encoder model on pooled multilingual data (Stage I). \textbf{Track "a"}: We fine-tune the multilingual encoder on the monolingual data for each language (Stage II) and finally freeze the CTC encoder to train a monolingual Contextual Adapter using the outputs from the (presumably) strongest possible encoder model (Stage III). In this track, we address the issue of the weak CTC encoder model. However, the training of the Contextual Adapters is still restricted to  a small dataset from a low-resource language. \textbf{Track "b"}: We design another track, where we freeze the CTC encoder model first and train the Contextual Adapters by pooling paired speech/text training examples and entity lists from
all languages (Stage II), and then fine-tune the full model (encoder and Contextual Adapters) on the monolingual data language for each language (Stage III). The details of this track are highlighted in Figure \ref{fig:arch}. This track helps to mitigate both the issues with training of the Contextual Adapters simultaneously. The multilingual Contextual Adapters helps in bootstrapping the `copying` mechanism for a low-resource language, while the multilingual CTC model acts as a strong baseline.


\section{Experiments}
\subsection{Data}
We evaluate our methods on from five different languages - Portuguese (pt), Italian (it), Spanish (es), English (en) and French (fr). Similar to \cite{houston2023exploration}, our training data volume varies from 100 hours for Portuguese (low-resource) to 2000 hours for French (high-resource). Table \ref{tab:data_partition} highlights the volumes of training and validation dataset for each language. This data comes from a mix of conversational telephony, media, news and call-center, with a mix of various dialects, sampling rates and acoustic conditions. 
We use the test split of the publicly available Librivox MLS dataset \cite{Pratap2020} for evaluation. It contains an average 8.9 hours of paired audio and text for each language.
To showcase the personalization aspect of our models, we provide a custom list of entity words to the model and observe the change in recognition of those specific words.  Since we do not have an explicit list of entities curated for our test dataset, we follow the same strategy as \cite{dingliwal2023personalization} and randomly choose 500 OOV words
per language from references i.e. the words that never appear in training transcripts. Table \ref{tab:data_partition} provides an overview of the evaluation transcripts, indicating the total number of tokens, as well as the number of tokens corresponding to the OOVs that are chosen as our entity lists. Although these words are only a small fraction of the total number of tokens in the transcripts, they are typically the most useful entity terms for the downstream application. Therefore, along with WERs, we track the recognition of these words using F1 scores (\%) directly. The F1 score is determined using the frequency of the custom word in the reference and the hypothesis of the ASR model.
\begin{table}[th]
     \centering
     \begin{tabular}{ l|cc||ccc }
       \toprule
       \textbf{Language} & \textbf{Train} & \textbf{Dev} & \multicolumn{3}{c}{\textbf{Test}}\\
        & \textbf{Hrs} & \textbf{Hrs} & \textbf{Hrs} & \textbf{\#Tokens} & \textbf{\#OOVs} \\
       \hline
       Portuguese & 100 & 71 & 3.7 & 31255 & 661 \\
       Italian & 500 & 64 & 5.3 & 40847 & 613 \\
       Spanish & 1000 & 35 & 10 & 88499 & 547 \\
       English & 1000 & 60 & 15.5 & 146747 & 693 \\
       French & 2000 & 61 & 10.1 & 93167 & 625 \\
       \hline
       Total & 4600 & 260 & 44.6 & 400515 & 3139 \\
       \bottomrule
     \end{tabular}
     \caption{Details of our datasets per language.}
     \vspace{-0.2cm}
     \label{tab:data_partition}
   \end{table}
\vspace{-0.8cm}
\begin{table*}[!htbp]
    \centering
    \begin{tabular}{l|l|cc|cc|cc|cc|cc}
         Model ID & Language (Train Hrs)  & \multicolumn{2}{c|}{\footnotesize{Portuguese (100h)}} & \multicolumn{2}{c|}{\footnotesize{Italian (500h)}} & \multicolumn{2}{c|}{\footnotesize{Spanish (1000h)}} & \multicolumn{2}{c|}{\footnotesize{English (1000h)}} & \multicolumn{2}{c|}{\footnotesize{French (2000h)}} \\
         & Experiment & \textbf{WER} & \textbf{F1} & \textbf{WER} & \textbf{F1} & \textbf{WER} & \textbf{F1} & \textbf{WER} & \textbf{F1} & \textbf{WER} & \textbf{F1} \\
         \toprule
         \multicolumn{12}{l}{\textbf{Monolingual}} \\
         \toprule
         \vspace{1.5pt} 
         MONO-I & Monolingual encoder & 47.1 & 32.4 & 21.8 & 48.4 & 8.2 & 67.9 & 21.2 & 20.6 & 9.7 & 46.4 \\
        \vspace{1.5pt} 
         MONO-II & \ \ +  Monolingual Adapters & 47.1 & 32.4 & 21.8 & 48.4 & 8.2 & 67.6 & 21.2 & 20.6 & 9.8 & 55.1 \\
        \vspace{1.5pt}   
         MONO-II.ce & \ \ \ \ \ \ + CE loss & 46.9  & 35.9 & 21.7 & 55.0 & 8.2 & 73.4 & 21.5 & 24.0 & 9.9 & 52.8 \\ 
         \toprule
         \multicolumn{12}{l}{\textbf{Multilingual Training (Track "a")}} \\
         \toprule
         \vspace{1.5pt} 
         ML-I & Multilingual encoder  & 27.5 & 64.4 & 20.9 & 54.4 & 10 & 69.4 & 23.7 & 21.1 & 10.7 & 45.4 \\
         \vspace{1.5pt} 
         ML-II.a & \ \ + Fine-tune encoder  & 22.0 & 70.7 & 19.7 & 52.1 & 7.7 & 70.4 & 19.1 & 20.3 & 8.7 & 48.6 \\
         \vspace{1.5pt} 
         ML-III.a & \ \ \ \ + Monolingual Adapters & 22.0 & 70.6 & 19.7 & 52.1 & 7.9 & 78.2 & 19.4 & 36.9 & 8.9 & 63.3 \\
         \vspace{1.5pt} 
         ML-III.a.ce & \ \ \ \ \ \ + CE loss & 22.1 & 71.5 & 19.7 & 54.3 & 7.8 & 74.7 & 19.3 & 28.4 & 8.9 & 57.1 \\
         \toprule
         \multicolumn{12}{l}{\textbf{Multilingual Contextual Adapters (Track "b")}} \\
         \toprule
         \vspace{1.5pt} 
         ML-I & Multilingual encoder  & 27.5 & 64.4 & 20.9 & 54.4 & 10 & 69.4 & 23.7 & 21.1 & 10.7 & 45.4 \\
         \vspace{1.5pt} 
         ML-II.b & \ \ + Multilingual Adapters  & 27.7 & 71.4 & 21.4 & 60.4 & 10.3 & 74.0 & 24.1 & 32.5 & 10.9 & 54.7 \\
        \vspace{1.5pt} 
        ML-II.b.ce & \ \ \ \ + CE Loss & 27.6 & 69.7 & 21.2 & 58.8 & 10.2 & 73.8 & 24 & 30.1 & 10.9 & 50.8 \\
         \vspace{1.5pt} 
        ML-III.b & \ \ \ \  + Fine-tune full model &  \textbf{19.3} & \textbf{80.7} & \textbf{18.2} & \textbf{73.0} & \textbf{7.5} & \textbf{83.8} & 18.4 & \textbf{49.3} & \textbf{8.5} & \textbf{70.0} \\
        ML-III.b.inf & \ \ \ \  - Adapters at inference & 19.5 & 71.2 & 18.3 & 50.0 & 7.5 & 69.0 & \textbf{18.2} & 20.7 & 8.6 & 43.9 \\

        \bottomrule
         
    \end{tabular}
    \vspace{1.5pt} 
    \caption{WER and F1 score (\%) on custom words for each language for monolingual, multilingual (Track "a" and "b") models.}
    \label{tab:f1}
    \vspace{-0.8cm}
\end{table*}

         

\subsection{Experimental Setting}
We use 12-layers of Conformer blocks with 8 self-attention heads, a 1024-dimensional feedforward layer and an input/output size of 384 for our CTC models \cite{gulati2020conformer}. These models directly predict subword targets generated from a common sentence-piece model for all languages with total vocabulary size of 2048. All our base CTC models are trained for 60 epochs with a maximum learning rate of 3e-3 after 20k warmup steps. We use greedy decoding algorithm for inference. 
The architecture and hyper-parameters for our Contextual Adapters resembles \cite{dingliwal2023personalization}. The total number of trainable parameters in the CTC models and the Contextual Adapters are 42 million and 1.25 million respectively. 
Since we do not have entities marked in our training data, we choose $w^{boost}$ to be the word in the utterance with the lowest frequency in the entire training text. 
$W^{boost}$ is a random subset of the combined list of such words from all the utterances. Similar to \cite{dingliwal2023personalization}, we use curriculum learning for training the Contextual Adapters by increasing the size of $W^{boost}$ from 30 to 100 over different epochs.
For our models with the additional CE loss ($\mathcal{L}_{CE}(.)$), we choose $\alpha=25$ based on visual inspection of training loss curves. The ESPnet library \cite{watanabe2018espnet} was used for all our implementations.

\section{Results}

\subsection{Monolingual Training}
\noindent \textbf{Cross Entropy loss is essential for training Contextual Adapters with limited data}:
The first set of rows in the Table \ref{tab:f1} summarizes the results of monolingual model training with Contextual Adapters for each of the five languages. For all of the languages except French, the conventional two-stage training of the model (encoder training followed by freezing encoder and training Contextual Adapters with CTC loss) does not improve the F1 score on the lists of custom words (Model IDs: MONO-I and MONO-II). However, adding the CE loss (Sec. \ref{subsec:adapter_loss}) for the training of the Contextual Adapters results in small F1 score improvements for all languages (Model ID: MONO-II.ce). This suggests that the guidance from the CTC loss alone is insufficient for the Contextual Adapters to learn useful representations in a limited data setting, whereas the proposed CE loss can encourage boosting the right word for the right set of time frames. However, all the improvements for custom entity recognition are modest for monolingual models, likely due to the weak base CTC model that produces encoder layer outputs that are not very useful to identify and boost the right custom word. 

\subsection{Multilingual Training (Track "a")}
In the next set of rows in the Table \ref{tab:f1}, we compare the performance of Contextual Adapters trained on a strong multilingual CTC encoder baseline. This encoder model is trained using pooled data from all five languages (Stage I). Although simple, this approach results in strong WER improvements for low-resource languages like Portuguese over baseline monolingual models (Model ID: MONO-I and ML-I). This is attributable to the known phenomenon of knowledge transfer from higher-resource to lower-resource languages during multilingual training. Also, as expected, WERs for higher-resource languages (Spanish, English and French) see little to no improvements in WER, or even a slight degradation. 
In an attempt to train presumably the strongest possible CTC encoder model, we further fine-tune the encoder on monolingual data from each language (Track "a", Stage II). As expected, this improve the WER for all languages over the pooled model (Model IDs: ML-I and ML-II.a). 
After fine-tuning the encoder model for all the languages, we train Contextual Adapters using monolingual data by freezing the fine-tuned encoder (Track "a", stage III). 
However, we still do not see any improvement in the custom entity recognition for low-resource languages like Portuguese and Italian (Model IDs: ML-II.a and ML-III.a). But we see that the F1 scores for entity recognition increases by 15\% with the Contextual Adapters for other high-resource languages like English and French. This suggests that conventional methods for training Contextual Adapters need sufficient data to learn the 'copying mechanism'. Similar to our observation for monolingual models, adding the CE loss during training of the Contextual Adapters provides minor improvements in the custom entity F1 scores for low-resource languages (Model ID: ML-III.a.ce). 

\subsection{Multilingual Contextual Adapters (Track "b")}
In the second track, we freeze the multilingual encoder and train multilingual Contextual Adapters (Track "b", Stage II) by pooling paired speech/text training examples and entity lists from all languages. As opposed to the track "a" approach, this results in an improvement in the F1 score of custom words for all languages, even for low-resource languages like Portuguese (Table \ref{tab:f1}, Model IDs: ML-I and ML-II.b). This suggests that Contextual Adapters for low-resource languages can benefit from the data of other languages to learn the `copying mechanism`. Here, we also observe that the Contextual Adapters bring similar performance improvements with and without the proposed CE loss (Model ID: ML-II.b and ML-II.b.ce). This suggests that while the CE loss is essential for training of the Contextual Adapters with limited data, it is unnecessary when sufficient data are available.  

\noindent \textbf{Contextual Adapters reduce the WER of the base CTC model}: 
Next, we fine-tune both the encoder and the Contextual Adapters from ML-II.b together on monolingual data from each language (Track "b", Stage III). We first highlight the effect of this approach on WER during inference. We find that fine-tuning both the multilingual encoder and the multilingual Contextual Adapters together significantly reduces WER over fine-tuning just the multilingual encoder alone. For low-resource Portuguese, we see a 11\% WER reduction, demonstrating that along with personalization, Contextual Adapters also aid in effective fine-tuning of the base encoder model. 

To determine whether this improvement can be attributed to the improved recognition of rare words by the Contextual Adapters, we also run inference using the jointly fine-tuned encoder alone (Model ID: ML-III.b.inf). We find an expected decrease in F1 scores, since custom entities are no longer being boosted, but overall WER remains largely unchanged.

This improved WER after joint fine-tuning can be attributed to the supplementary training objective of Contextual Adapters, which involves distinguishing a target entity term embedded in the audio signal from a pool of random entity terms. This task, in conjugation with prediction of the correct sub-word token sequence, guides the base CTC encoder model to learn more meaningful audio representations. Moreover, our implementation of Contextual Adapters utilizes a weighted combination of features extracted from all Conformer blocks within the CTC encoder, allowing it have a more direct impact on the gradients used to update all layers than CTC loss, which only receives outputs from the final softmax layer. We plan to investigate the benefits of training Contextual Adapters to improve the base encoder model further as part of a future work. 

\noindent \textbf{Multilingual Contextual Adapters achieve the best custom entity recognition}:  
Apart from achieving the best WER for all the languages (Model ID: ML-III.b), this model also performs the best for custom entity word recognition. Joint fine-tuning results adds improvements over multilingual encoder and Contextual Adapter training only (Model ID: ML-II.b) between ~10-20\%. For low-resource Portuguese, the F1 score for OOV word recognition is 48\% better than the conventional monolingual Contextual Adapters. Even for languages like French with a relatively large amount of training data, there is a substantial improvement in F1 scores from multilingual Contextual Adapters training compared to just monolingual training (55.1\% and 70.0\%, respectively).
The disparity in F1-scores between the two tracks of our multilingual training demonstrates the importance of multilingual data in training Contextual Adapters, and suggests that the copying mechanism it learns is language-agnostic (at least for this set of languages).



\vspace{-0.4cm}
\section{Conclusion}
\vspace{-0.2cm}
In this work, we demonstrated that approaches such as Contextual Adapters \cite{dingliwal2023personalization}, which aim to bias CTC models towards a specific set of words, are ineffective in languages with scarce data.The main factors contributing to this behavior are the complex nature of the training objective, a deficient CTC base model, and insufficient data diversity to facilitate the learning of the boosting mechanism.To mitigate these issues, we first propose a multi-task cross entropy loss to train the Contextual Adapters. Next, we investigate the pooling of multilingual data as a means of transferring knowledge about the boosting mechanism from a high-resource language to a low-resource one. By utilizing our multilingual Contextual Adapters, we were able to achieve superior custom entity recognition, resulting in a 48\% increase in F1 score for retrieving these entities, as opposed to a mere 7\% improvement achievable with baseline methods. Interestingly, we observe that jointly training the Contextual Adapters and the base CTC model together reduces WER  of the base CTC model by 5-11\% as a by-product, even when the adapters are not used during inference. 

\pagebreak



\bibliographystyle{IEEEtran}
\bibliography{mybib}

\end{document}